\documentclass[10pt,twocolumn,letterpaper]{article}

\usepackage[pagenumbers]{arxiv} 

\usepackage{graphicx}
\usepackage{amsmath}
\usepackage{amssymb}
\usepackage{booktabs}
\usepackage{pdfpages}

\usepackage[pagebackref,breaklinks,colorlinks]{hyperref}

\usepackage[capitalize]{cleveref}
\crefname{section}{Sec.}{Secs.}
\Crefname{section}{Section}{Sections}
\Crefname{table}{Table}{Tables}
\crefname{table}{Tab.}{Tabs.}

\usepackage{xcolor}
\usepackage{xspace}
\usepackage{diagbox}

\usepackage[utf8]{inputenc}
\usepackage{adjustbox}
\usepackage{wrapfig}
\usepackage{multirow}
\newcommand*\rot{\rotatebox{90}}

\usepackage{graphicx}
\usepackage{sidecap}
\usepackage{caption}
\usepackage{subcaption}
\usepackage{booktabs}
\usepackage{amsfonts}
\usepackage{amsmath, bm}
\usepackage{mathtools}
\DeclarePairedDelimiter\ceil{\lceil}{\rceil}

\usepackage{xcolor, colortbl}
\definecolor{grey}{rgb}{0.9, 0.9, 0.9}

\definecolor{ffe1da}{RGB}{255,225,218}
\definecolor{F7E0D5}{RGB}{247,224,213}
\definecolor{darkF7E0D5}{RGB}{209,154,128}
\definecolor{Green}{RGB}{0,204,0}
\colorlet{Light}{F7E0D5}

\usepackage{algorithm}
\usepackage{listings}
\usepackage{textpos}
\usepackage{comment}
\usepackage{etoolbox}
\makeatletter
\AfterEndEnvironment{algorithm}{\let\@algcomment\relax}
\AtEndEnvironment{algorithm}{\kern2pt\hrule\relax\vskip3pt\@algcomment}
\let\@algcomment\relax
\newcommand\algcomment[1]{\def\@algcomment{\footnotesize#1}}
\renewcommand\fs@ruled{\def\@fs@cfont{\bfseries}\let\@fs@capt\floatc@ruled
  \def\@fs@pre{\hrule height.8pt depth0pt \kern2pt}%
  \def\@fs@post{}%
  \def\@fs@mid{\kern2pt\hrule\kern2pt}%
  \let\@fs@iftopcapt\iftrue}
\makeatother

\makeatletter
\DeclareRobustCommand\onedot{\futurelet\@let@token\@onedot}
\def\@onedot{\ifx\@let@token.\else.\null\fi\xspace}

\def\ie{\emph{i.e}\onedot}

\makeatother

\newcommand{\cP}{\mathcal{P}}

\newcommand{\cX}{\mathcal{X}}

\def\cvprPaperID{***} 
\def\confName{CVPR}

\begin{document}

\title{Group Generalized Mean Pooling for Vision Transformer}

\author{Byungsoo Ko$^1$,\space\space
Han-Gyu Kim$^2$,\space\space
Byeongho Heo$^3$,\\
Sangdoo Yun$^3$,\space\space
Sanghyuk Chun$^3$,\space\space
Geonmo Gu$^1$,\space\space
Wonjae Kim$^3$\\
{\tt\small $^1$NAVER Vision,\space\space $^2$NAVER Clova Speech,\space\space $^3$NAVER AI Lab}\\
}

\maketitle

\begin{abstract}
Vision Transformer (ViT) extracts the final representation from either class token or an average of all patch tokens, following the architecture of Transformer in Natural Language Processing (NLP) or Convolutional Neural Networks (CNNs) in computer vision.
However, studies for the best way of aggregating the patch tokens are still limited to average pooling, while widely-used pooling strategies, such as max and GeM pooling, can be considered.
Despite their effectiveness, the existing pooling strategies do not consider the architecture of ViT and the channel-wise difference in the activation maps, aggregating the crucial and trivial channels with the same importance.
In this paper, we present Group Generalized Mean (GGeM) pooling as a simple yet powerful pooling strategy for ViT.
GGeM divides the channels into groups and computes GeM pooling with a shared pooling parameter per group.
As ViT groups the channels via a multi-head attention mechanism, grouping the channels by GGeM leads to lower head-wise dependence while amplifying important channels on the activation maps.
Exploiting GGeM shows 0.1\%p to 0.7\%p performance boosts compared to the baselines and achieves state-of-the-art performance for ViT-Base and ViT-Large models in ImageNet-1K classification task.
Moreover, GGeM outperforms the existing pooling strategies on image retrieval and multi-modal representation learning tasks, demonstrating the superiority of GGeM for a variety of tasks.
GGeM is a simple algorithm in that only a few lines of code are necessary for implementation.

\end{abstract}
\section{Introduction}
\label{sec:introduction}

Over the past several years, there has been a breakthrough of Transformer networks in the Natural Language Processing (NLP) domain~\cite{devlin2019bert,brown2020language}.
It brought great interests in the computer vision community to exploit the Transformer architecture for vision tasks.
As a result, Vision Transformer (ViT)~\cite{dosovitskiy2021an} was introduced, and its variants have shown great success in image recognition~\cite{touvron2021deit,liu2021swin,heo2021pit,wang2021pvt}, self-supervised learning~\cite{caron2021dino,he2021masked,zhou2021ibot,baevski2022data2vec,bao2021beit,xie2021simmim}, object detection~\cite{li2022exploring,carion2020end,fang2021you,song2021vidt,song2022vidtplus}, segmentation~\cite{strudel2021segmenter,chen2021transunet}, image compression~\cite{kim2021joint}, image retrieval~\cite{el2021training,dubey2021vision}, and multi-modal representation learning~\cite{kim2021vilt,radford2021clip,mu2021slip}.
Compared to the previous standard vision models, such as Convolutional Neural Networks (CNNs), recent works have demonstrated that ViT shows equal or superior performance in image recognition tasks when equipped with huge number of parameters or large-scale datasets~\cite{dosovitskiy2021an,raghu2021vision}.

The ViT at early stage followed conventional architecture of Transformers used for NLP, letting the final state of an additional class token~\cite{devlin2019bert} be the feature representation.
However, using the class token as a feature representation can ignore per-token information.
Thus, there has been active studies on pooling strategies in NLP, aggregating per-token information as an alternative of the class token~\cite{li2020sentence,reimers2019sentence,gao2021simcse}.
For ViT, recent works exploit the average pooling for achieving better performances than the class token~\cite{park2022vision,pan2021scalable,chu2021conditional}, or preserving such per-token information~\cite{he2021masked,baevski2022data2vec,bao2021beit,xie2021simmim}.
Alternatively, we can explore the representative pooling strategies studied in CNNs, which are reported to be effective: \ie, max~\cite{razavian2016visual,Tolias2016rmac} and Generalized Mean (GeM)~\cite{Radenovic2018fine} poolings.
However, they have not been built with consideration of ViT architecture and do not consider channel-wise differences in the activation maps, aggregating crucial and trivial channels with the same importance.

In this paper, we present Group Generalized Mean (GGeM) pooling as a simple yet powerful pooling strategy for ViT.
GGeM divides channels into multiple groups in the final activation maps and computes GeM pooling with a shared pooling parameter per group (illustrated in Fig.~\ref{fig:comparison}).
As shown in Fig.~\ref{fig:feature_map}, each channel in the final activation maps activates a different spatial area.
GGeM considers such channel-wise differences by exploiting different trainable parameters for each group.
Moreover, we have analyzed how pooling strategies work for ViT and discover the followings: 1) the pooling parameters decide the degree of concentration of gradient, 2) the grouping in GGeM leads to a lower inter-head similarity, and 3) the higher pooling parameter increases the number of heads focusing on global information.
Compared to the existing pooling strategies, GGeM shows a 0.1\%p to 0.7\%p performance boost and achieves a new state-of-the-art performance for ViT-Base and ViT-Large on ImageNet-1K classification tasks.
Additionally, experiments on image retrieval and multi-modal representation learning demonstrate the versatility of GGeM for various vision tasks.
\section{Related Work}
\label{sec:related}

\paragraph{Vision Transformer and Self-supervised Learning.}
Inspired by its great success in NLP tasks, ViT~\cite{dosovitskiy2021an} has been introduced by following the conventional architecture of NLP Transformer, using a class token for final feature representation.
Recently, a number of studies~\cite{touvron2021deit,liu2021swin,heo2021pit,wang2021pvt,chen2021crossvit} have explored to improve ViT in terms of recognition performance and training efficiency.
DeiT~\cite{touvron2021deit} exploits the distillation scheme for better training efficiency, using both class and distillation tokens for feature representation.
Moreover, ViT has been actively used for Self-Supervised Learning (SSL) task~\cite{caron2021dino,zhou2021ibot,wei2021masked,he2021masked,xie2021simmim,bao2021beit,baevski2022data2vec}.
Recent ViT based SSL approaches, such as MAE~\cite{he2021masked}, SimMIM~\cite{xie2021simmim}, BeiT~\cite{bao2021beit} and Data2Vec~\cite{baevski2022data2vec}, feed direct loss to patch tokens for each objective (\ie, pixel-wise reconstruction or masked image modeling).
As the patch token contains rich spatial information, those SSL methods use the average pooling of all patch tokens to aggregate the per-token information.
Our goal is to propose a generic pooling method for both scratch training and fine-tuning.

\paragraph{Pooling Strategies}
have been widely studied to design CNN-based global descriptors in image retrieval tasks~\cite{spoc,razavian2016visual,Tolias2016rmac,jun2019combination,li2017ms,wu2018weighted}.
Popular pooling strategies in CNNs include the average pooling (a.k.a. SPoC)~\cite{spoc} and max pooling (a.k.a. MAC)~\cite{razavian2016visual,Tolias2016rmac}, which average and select maximum activations on the feature map, respectively.
GeM~\cite{Radenovic2018fine} has been introduced to generalize max and average pooling by a pooling parameter.
As variants of such standard pooling strategies, weighted sum pooling~\cite{kalantidis2016cross}, regional MAC~\cite{Tolias2016rmac}, multiscale RMAC~\cite{li2017ms}, and weighed GeM~\cite{wu2018weighted} have been introduced.
There also has been an attempt to use the attention mechanism for pooling by replacing the average pooling in CNNs with a single layer of multi-head attention block~\cite{radford2021clip}.
However, there are limited studies on pooling strategies w.r.t. activation maps (patch tokens) in ViT.
Thus, this work covers exploring how standard pooling strategies work in ViT.

\paragraph{Group-wise Computation.}
The concept of groups has been widely studied for CNN.
Group convolution has been proposed in AlexNet~\cite{krizhevsky2012alexnet} to distribute the model over two GPUs.
ResNeXt~\cite{xie2017aggregated} presents a module that splits channel dimensions into groups as group convolution for better performance under similar computational costs.
MobileNet~\cite{howard2017mobilenets} and Xception~\cite{chollet2017xception} adopt depth-wise separable convolutions, which are group convolutions with the same group number as the channel number.
Group Normalization~\cite{wu2018group} divides the channels into groups and computes the mean and variance within each group for normalization.
Likewise, the concept of the grouping can be found in Transformer.
In a Transformer block, the channels of intermediate representations are divided by the number of heads (\ie, channel-wise grouping) and computes attention within each head.
GGeM shares the same spirit of dividing channels into groups, but it performs GeM pooling within each group for effective aggregation of per-token information.

\section{Method}
\label{sec:method}

In this section, we propose GGeM, a pooling method for considering channel-wise differences of the activation maps by grouping channels.
We first formulate ViT architecture (Sec.~\ref{sec:vit}) and revisit representative pooling strategies studied in CNNs (Sec.~\ref{sec:pooling}).
Next, we present the motivation and details of GGeM (Sec.~\ref{sec:ggem}), and analyze how pooling strategies work and affect ViT (Sec.~\ref{sec:analysis}).

\begin{figure}[t]
    \centering
    \includegraphics[width=1.0\linewidth]{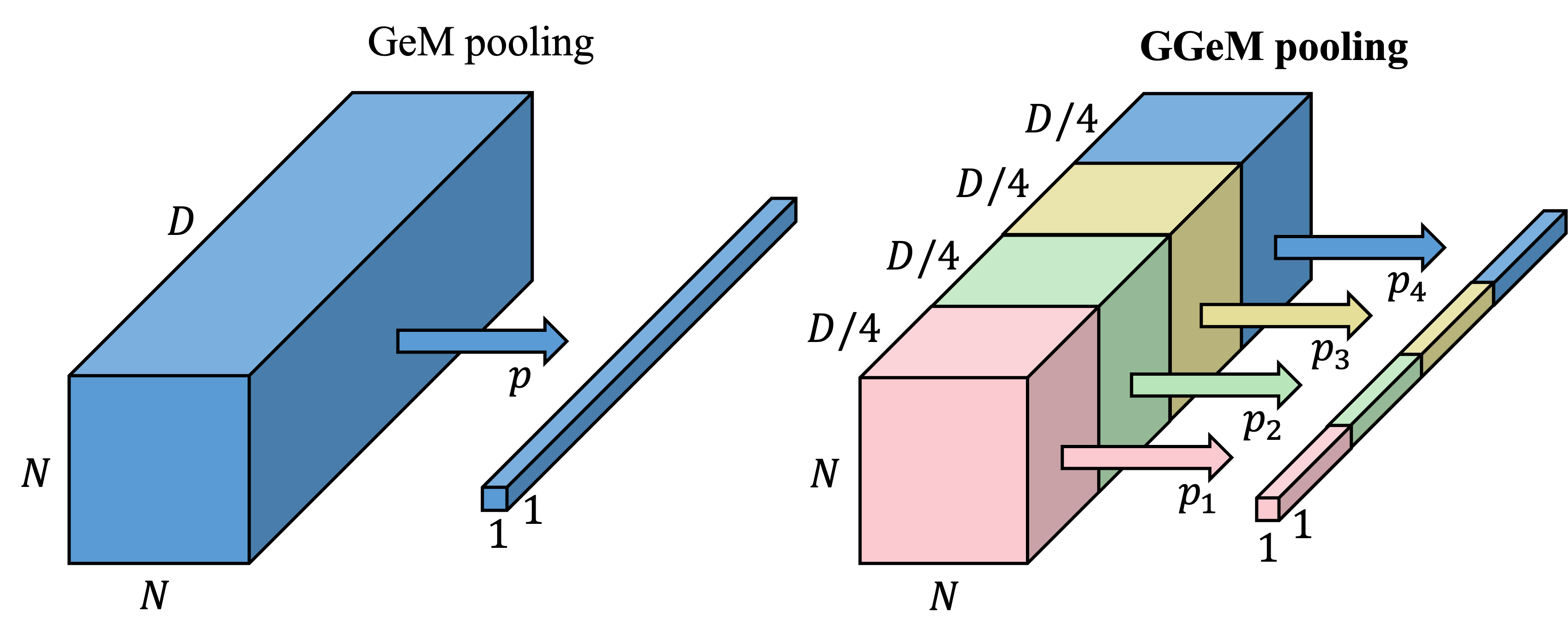}
    \caption{\textbf{Comparison between GeM and GGeM pooling.} Given $N \times N \times D$ activation maps, GeM aggregates them with a single pooling parameter $p$, while GGeM performs group-wise aggregation with different pooling parameter $p_i$ for each group. In this case, the number of group $G$ is 4.\vspace{-0.3cm}}
    \label{fig:comparison}
\end{figure}

\subsection{Vision Transformer}
\label{sec:vit}
This paper considers Vision Transformer models, such as ViT-Small, ViT-Base, or ViT-Large, for pooling strategies.
Given an image $I \in \mathbb{R}^{H \times W \times C}$, the image tensor is reshaped into a sequence of flattened 2D patches $I' \in \mathbb{R}^{N^2 \times ({R}^2C)}$, where $(H, W)$ is the resolution of the input image, $C$ is the number of channels, $(R, R)$ is the resolution of each image patch, and $(N, N)$ is the number of patches for height and width\footnote{The original ViT~\cite{dosovitskiy2021an} uses $I' \in \mathbb{R}^{(NM) \times ({R}^2C)}$, where $(NM)=HW/R^2$ to allow non-square images. In this paper, we assume square images for simplicity.}.
Then, the sequence of patches maps to $D$ dimensional feature by a trainable linear projection, and a learnable embedding is attached to the sequence of embedded patches as BERT's $[class]$ token.
Then, the embeddings are fed into the multiple Transformer blocks, and the last Transformer block outputs $\mathbb{R}^{({N}^2 + 1) \times D}$, where 1 is for the class token.
Usually, the $1 \times D$ class token is used for the final feature representation~\cite{dosovitskiy2021an,touvron2021deit}.
However, selecting the class token as the final representation loses the per-token information of the remaining $N^2$ token embeddings.
Hence, we are interested in how to aggregate the per-token information of the $\mathbb{R}^{N \times N \times D}$ patch tokens.

\begin{figure}[t]
    \centering
    \includegraphics[width=0.9\linewidth]{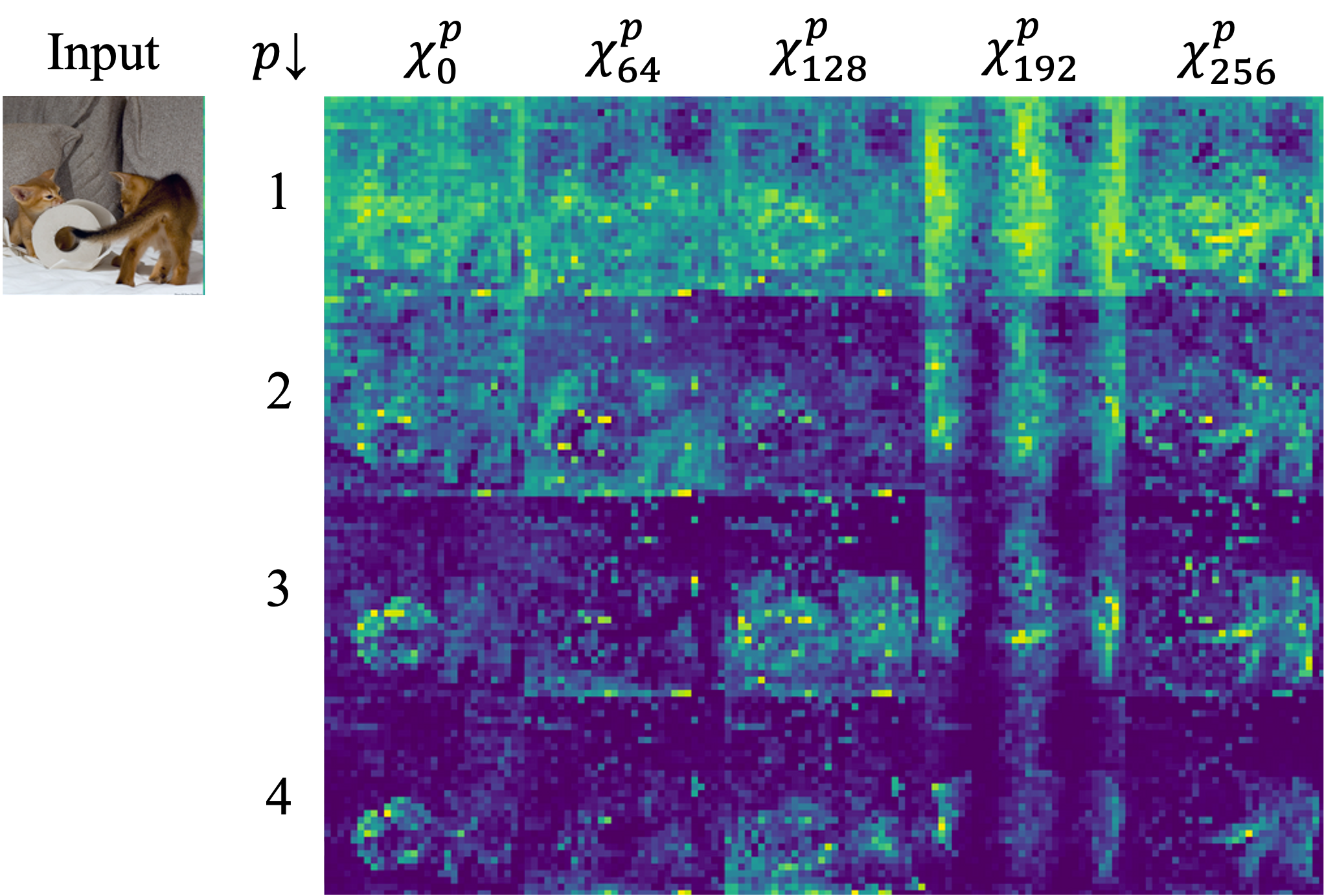}
    \caption{\textbf{Visualization of the activation map $\cX_d^p$.} We use ImageNet-1K trained models using GeM pooling with fixed $p$. Each row indicates pooling parameter $p$, where $p=1$ equals to average pooling. Each column indicates the index of random channel dimension $d$.}
    \label{fig:feature_map}
\end{figure}

\subsection{Pooling Strategies}
\label{sec:pooling}
\paragraph{Average Pooling.}
We denote $\cX \in \mathbb{R}^{N \times N \times D}$ as a 3D tensor of activation maps and $\cX_d$ as the set of $N \times N$ activations of a feature map $d \in \{1 \cdots D\}$.
Given the activation maps $\cX$, we aggregate the spatial information into a feature representation $\bm{v} \in \mathbb{R}^{D}$, an output of a pooling process.
The average pooling~\cite{spoc} has been a standard pooling strategy for CNNs and also has been actively exploited in ViT~\cite{he2021masked,bao2021beit,baevski2022data2vec,xie2021simmim}.
Let $\mid \cX_d \mid$ be the number of elements in the set $\cX_d$, then the feature representation in the case of the conventional average pooling is as follows:
\begin{equation}
    \bm{v}^{(a)}=[v_1^{(a)},\cdots,v_d^{(a)},\cdots,v_D^{(a)}]^T, \quad v_d^{(a)} = \frac{1}{\mid \cX_d \mid}\sum_{x \in \cX_d}{x}.
    \label{eq:average}
\end{equation}

\begin{figure}[t]
    \centering
    \includegraphics[width=0.75\linewidth]{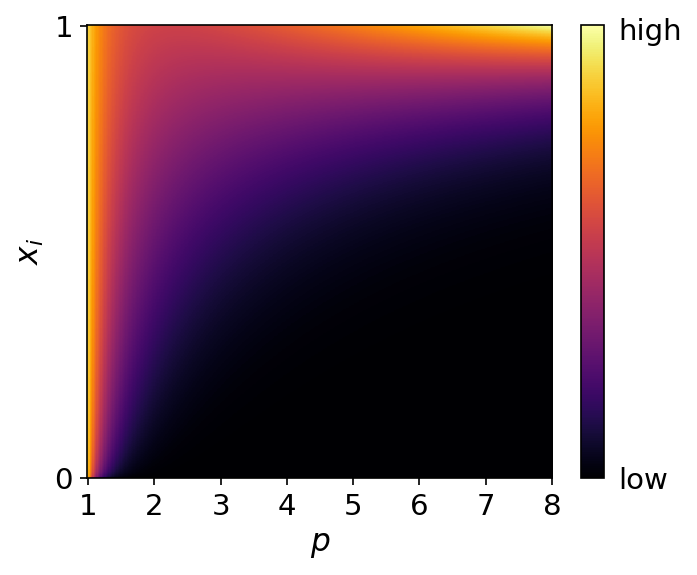}
    \caption{\textbf{The heatmap of partial derivatives ${\partial v_d^{(g)}} / {\partial x_i}$.} The feature before a pooling $x_i$ and pooling parameter $p$ are uniformly generated from 0 to 1 and 1 to 8, respectively.}
    \label{fig:gradient_gem}
\end{figure}

\paragraph{Max Pooling.} Instead of taking the average of the feature map, the max pooling chooses the highest activation in the feature map to capture the most distinctive representation.
The feature representation obtained by the conventional global max pooling~\cite{razavian2016visual,Tolias2016rmac}, is defined by:
\begin{equation}
    \bm{v}^{(m)}=[v_1^{(m)},\cdots,v_d^{(m)},\cdots,v_D^{(m)}]^T, \quad v_d^{(m)} = \max_{x \in \cX_d}{x}.
    \label{eq:max}
\end{equation}

\paragraph{Generalized Mean Pooling.} By exploiting the generalized mean~\cite{dollar2009integral}, the Generalized Mean Pooling (GeM)~\cite{Radenovic2018fine} has been introduced as:
\begin{equation}
    \begin{split}
    \bm{v}^{(g)}=[v_1^{(g)},\cdots,v_d^{(g)},\cdots,v_D^{(g)}]^T,\\
    \quad v_d^{(g)} = \left(\frac{1}{\mid \cX_d \mid}\sum_{x \in \cX_d}{x^{p_d}}\right)^\frac{1}{p_d},
    \label{eq:gem}
    \end{split}
\end{equation}
where pooling parameter $p_d$ is either trainable or fixed.
In practice, a shared pooling parameter $p$ is used for all channels as $p_d=p, \forall d \in \{1, \cdots, D\}$.
This is because using different pooling parameters per channel is reported to be distractive for training~\cite{Radenovic2018fine}.
Thus, the term ``GeM'' in this paper without extra explanation denotes the typical GeM that uses a shared pooling parameter $p$ for all channels.
The average pooling and the max pooling are special cases of GeM pooling: GeM becomes average pooling when $p_d=1$; GeM becomes max pooling when $p_d \rightarrow \infty$.

\subsection{Group Generalized Mean Pooling}
\label{sec:ggem}
The role of pooling parameter $p$ in GeM is to amplify the discrepancy of activations within the feature map.
We visualize randomly selected feature maps $\cX_d^p$ of ImageNet-1K~\cite{deng2009imagenet} trained models using GeM pooling with fixed pooling parameter $p$.
As shown in Fig.~\ref{fig:feature_map}, each channel dimension contains different spatial information.
Moreover, the feature map of the average pooling ($p=1$) shows activations in diverse regions, while it is shown that the larger $p$ becomes, the more locally activation map responses.
With these observations, our motivation is to stress important channels while suppressing trivial channels using different pooling parameters $p_d$ for channels.

We propose Group Generalized Mean pooling (GGeM).
GGeM divides the channels into groups and computes GeM with the pooling parameter $p_i$ per group.
Here, we assume that each group of channels is ordered sequentially along the $D$ axis.
The number of groups $G$ is a pre-defined hyper-parameter, where we suggest using the number of heads in ViT (\ie, 6 for ViT-S, 12 for ViT-B, and 16 for ViT-L) as $G$.
Let $\cP = \{p_1, \cdots, p_G\}$ be a set of trainable parameters $p_i$, the feature vector ($\bm{v}^{(gg)}$) obtained by GGeM is defined by:
\begin{equation}
    \begin{split}
    \bm{v}^{(gg)}=[v_1^{(gg)},\cdots,v_d^{(gg)},\cdots,v_D^{(gg)}]^T,\\
    v_d^{(gg)} = \left(\frac{1}{\mid \cX_d \mid}\sum_{x \in \cX_d}{x^{p_{y(d)}}}\right)^\frac{1}{p_{y(d)}}, \quad y(i) =  \ceil*{\frac{i}{D/G}},
    \label{eq:ggem}
    \end{split}
\end{equation}
where $\lceil \cdot \rceil$ is the ceiling operation and $D/G$ is the number of channels per group.
GGeM with $G=1$ is equivalent to GeM with a shared pooling parameter $p$ for all channels.
On the other hand, GGeM with $G=D$ contains $D$ pooling parameters $\{p_1,\cdots,p_D\}$, one for each channel.
The comparison between GeM and GGeM is illustrated in Fig.~\ref{fig:comparison}.

\subsection{Analysis}
\label{sec:analysis}

\begin{figure}[t]
    \centering
    \includegraphics[width=0.69\linewidth]{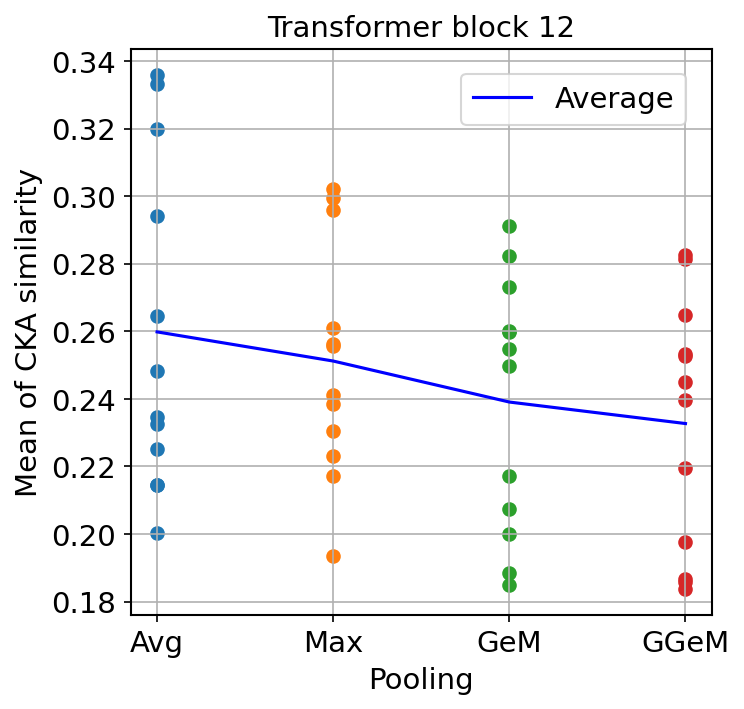}
    \caption{\textbf{Mean of CKA similarities between heads of MHA.} Each dot denotes mean of CKA similarities between the query head and other heads. The blue line denotes average of all CKA similarities between heads.}
    \label{fig:head_cka}
\end{figure}

\paragraph{Gradient Flow.} 
In order to find out the relationship between the feature before a pooling $x_i\, (i\in \{1, \cdots, N^2\})$ and the feature after a pooling $v_d^{(g)}$ during the back-propagation, we analyze the partial derivative of the pooled feature $v_d^{(g)}$ with respect to the input of pooling layer $x$ of Eq.~\ref{eq:gem}.
Such partial derivative ${\partial v_d^{(g)}} / {\partial x_i}$ is computed as follows~\cite{Radenovic2018fine}:
\begin{equation}
    \frac{\partial v_d^{(g)}}{\partial x_i} = \frac{1}{\mid \cX_d \mid} {v_d^{(g)}}^{1-p_d} x_i^{p_d - 1} .
    \label{eq:gradient_gem}
\end{equation}
According to Eq.~\ref{eq:gradient_gem}, because all inputs share the same $\mid \cX_d \mid$ and the same $v_d^{(g)}$, a larger input $x_i$ achieves a larger derivative ${\partial v_d^{(g)}} / {\partial x_i}$ compared to that of a smaller input.
Besides, such relative difference of the derivatives caused by input difference will become much larger with a larger $p_d$.
The growth of the relative difference is proportional to ${p_d - 1}$ power of the input difference because of the existence of the term $x_i^{p_d - 1}$ in Eq.~\ref{eq:gradient_gem}.
Such phenomenon implies that: for a smaller $p_d$, gradients would be evenly assigned to each input; for a larger $p_d$, more gradient is assigned to the input with a larger value.
In other words, \textit{the pooling parameter $p_d$ decides the degree of concentration of gradient}.
The same conclusion also can be found in Fig.~\ref{fig:gradient_gem}, where the heatmap of derivatives is provided with the input $x_i$ uniformly generated from 0 to 1, and $p$ uniformly generated from 1 to 8.

\begin{figure}[t]
    \centering
    \includegraphics[width=0.65\linewidth]{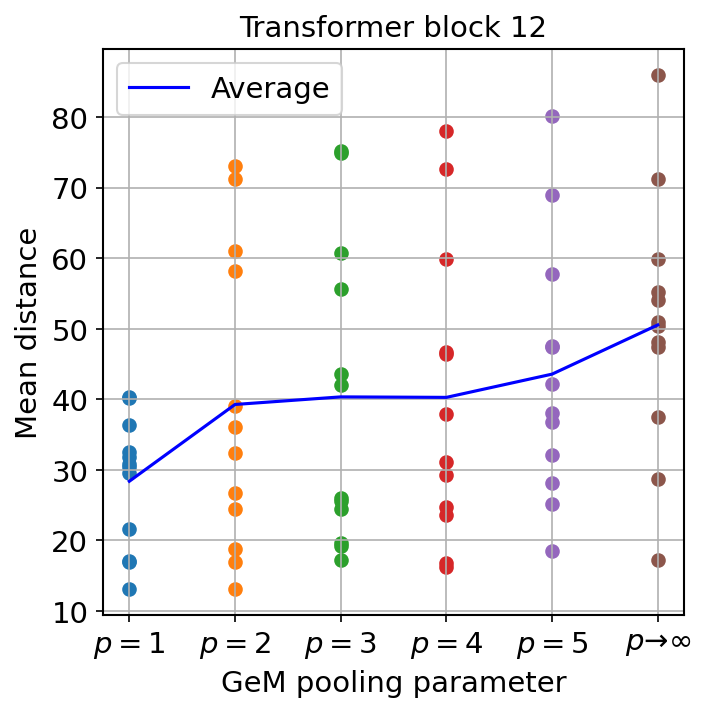}
    \caption{\textbf{Attention head mean distances with different GeM pooling parameters.} Higher mean distance indicates more globally attending heads. $p=1$ and $p \rightarrow \infty$ are average and max pooling, respectively.}
    \label{fig:head_dist}
\end{figure}

\paragraph{Impact of Grouping.}
For the well-trained Multi-Head Attention (MHA) layer of a transformer, the inter-head similarity will be small so that different heads cover different subspaces~\cite{caron2021dino,cordonnier2020mha,voita2019analyzing,michel2019sixteen}.
However, MHA does not always guarantee independently acting heads focusing on different subspaces~\cite{cordonnier2020mha}.
As all heads share the same $p$, average, max and GeM pooling can not guide each channel to have different characteristics, which is not beneficial for leading heads to make different subspaces.
However, GGeM, which uses different $p_i$ per head, is beneficial for inducing heads to learn characteristics of different subspaces because different heads can achieve gradients with different degrees of concentration.
In order to show such characteristics of GGeM, we have analyzed the similarity of heads by taking the average over 2,000 input images on ImageNet-1K trained ViT-B/16.
We use Centered Kernel Alignment (CKA) similarity~\cite{kornblith2019similarity} for the analysis, which is used to measure the similarity of different layers~\cite{raghu2021vision}.
As shown in Fig.~\ref{fig:head_cka}, GGeM, which uses different $p_d$ for different heads, shows smaller inter-head similarity compared to other pooling methods.
This implies that \textit{GGeM is beneficial for inducing the heads to learn different representation subspaces}.
Similar patterns stand out in the later Transformer blocks (\ie, Transformer blocks 10 to 12).
See the Appendix for the different blocks.

The weights of each head of MHA are tightly coupled as they are trained in a bundle.
However, for GGeM that uses independent $p_d$ for all dimensions (\ie, $G = D$), the different degrees of gradient concentration for the different dimensions of the head can disturb the ``training in a bundle'' process.
On the contrary, optimal GGeM with the number of groups equal to the number of heads uses one $p_i$ per head, which will keep the weights of each head to be trained in a bundle.
In conclusion, GGeM is more beneficial for performance improvement compared to other types of GeM methods in ViT.

\paragraph{Attention Head Mean Distance.}
In order to understand how pooling strategies affect ViT models, we analyze self-attention layers following~\cite{dosovitskiy2021an,raghu2021vision}.
Each self-attention layer consists of multiple self-attention heads as channel-wise grouping, and each head aggregates information from other spatial locations by attention mechanism.
Here, we compute the head mean distance between a query patch position and the other positions it attends to.
In detail, we weight the pixel distances with the attention weights for each attention head.
Then we average it per head over 2,000 input images by using ImageNet-1K trained ViT-B/16.
Interestingly, we see a clear pattern in the head mean distances as shown in Fig.~\ref{fig:head_dist}.
By increasing the pooling parameter $p$ from average ($p=1$) to max ($p \to \infty$) pooling, the average of all head mean distance increases, which indicates the number of heads attending on global information increases.
While average pooling more attends on local information compared to other pooling methods, \textit{the increase of $p$ diversifies the role of the heads attending on both local and global information}.
Similar patterns can be observed in the later Transformer blocks (\ie, Transformer block 10 to 12).
See the Appendix for the different blocks.

\begin{figure}[t]
    \centering
    \includegraphics[width=0.7\linewidth]{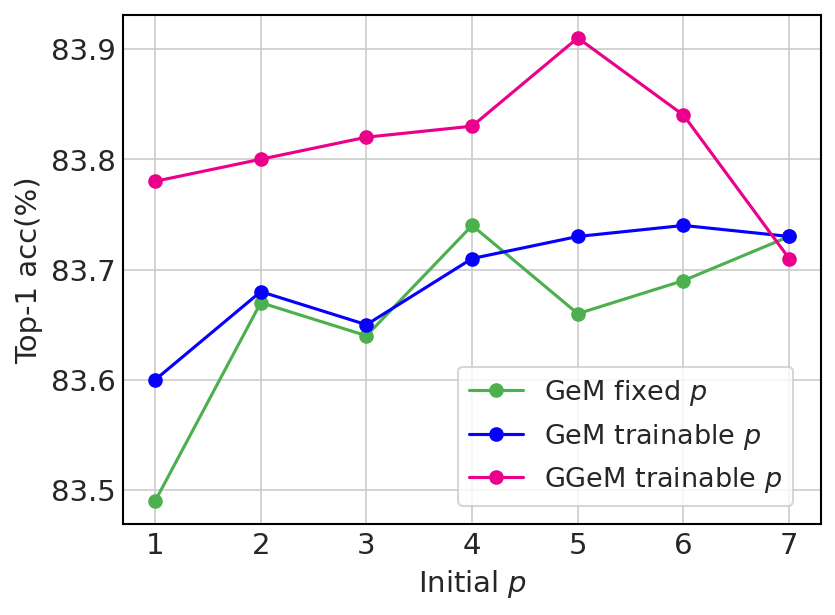}
    \caption{\textbf{Impact of initial $p$.} We differentiate initial $p$ for GeM with fixed $p$, GeM with trainable $p$ and GGeM with trainable $p$.}
    \label{fig:power}
\end{figure}

\begin{figure}[h]
    \centering
    \includegraphics[width=0.7\linewidth]{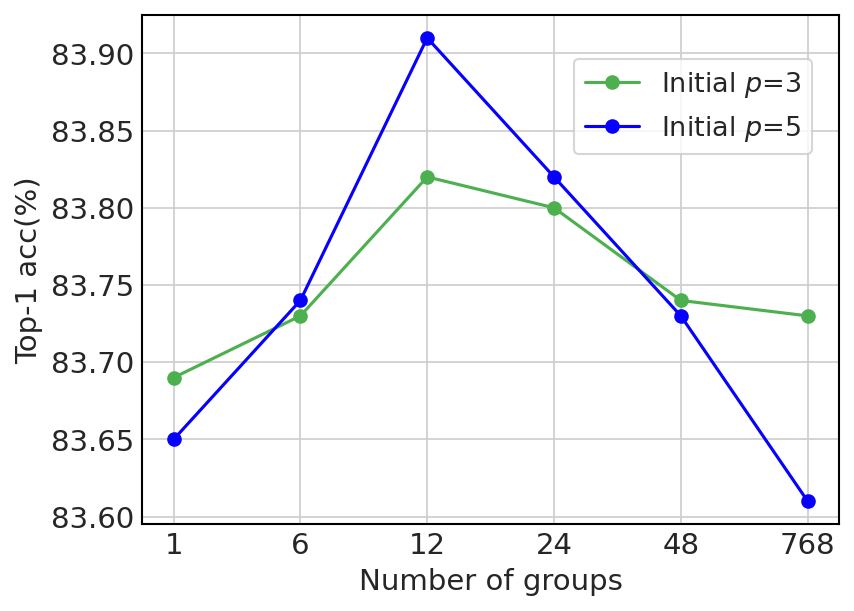}
    \caption{\textbf{Impact of number of groups.} $G=1$ is the GeM pooling, while $G=12$ corresponds to the number of heads in ViT-B/16.}
    \label{fig:group}
\end{figure}

\section{Experiments}
\label{sec:experiment}
In this section, we validate the proposed GGeM on different vision tasks.
We conduct ablation studies on the main properties and robustness of GGeM, and show image classification performance on scratch training, fine-tuning, linear probing and partial fine-tuning (Sec.~\ref{sec:classification}).
Next, we perform experiments on image retrieval (Sec.~\ref{sec:retrieval}) and multi-modal representation learning (Sec.~\ref{sec:multimodal}).

\subsection{Image Classification}
\label{sec:classification}
We perform an ablation study of GGeM and evaluate the image classification performance of different strategies on the ImageNet-1K~\cite{deng2009imagenet} dataset.
For training from scratch, we set the training epoch to 300 epochs.
For fine-tuning a self-supervised model, the training epoch is set to 100 epochs and 50 epochs for ViT-B/16 and ViT-L/16, respectively.
As a baseline model for the ablation study, we use MAE~\cite{he2021masked} pre-trained ViT-B/16 model unless otherwise noted in the experiment.
We report top-1 validation accuracy of a $224 \times 224$ cropped image.
Experimental details are in the Appendix.

\subsubsection{Main Properties}
\label{sec:properties}

\begin{figure*}[t]
    \centering
    \begin{subfigure}[c]{0.7\columnwidth}
        \centering
        \includegraphics[width=\columnwidth]{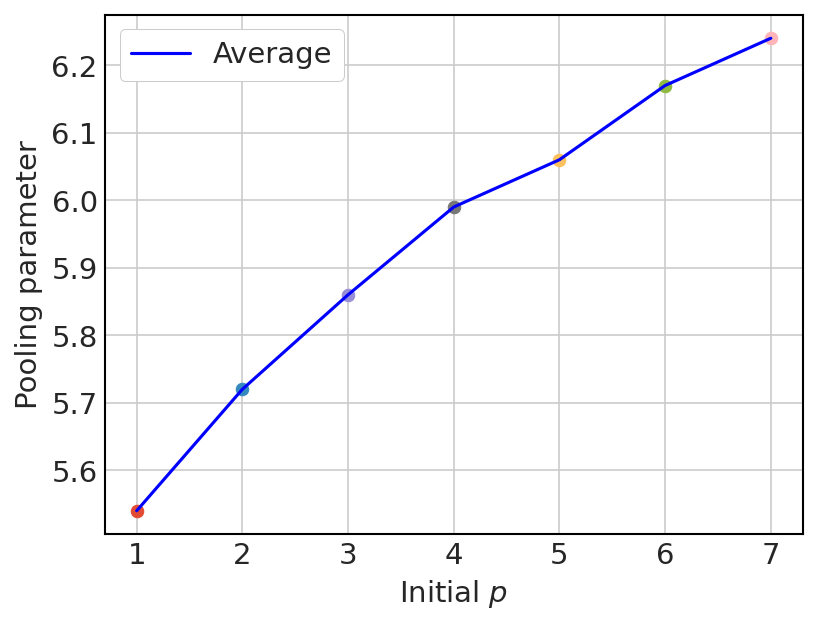}
        \caption{GeM pooling}
    \end{subfigure}
    \hspace{10mm}
    \begin{subfigure}[c]{0.7\columnwidth}
        \centering
        \includegraphics[width=\columnwidth]{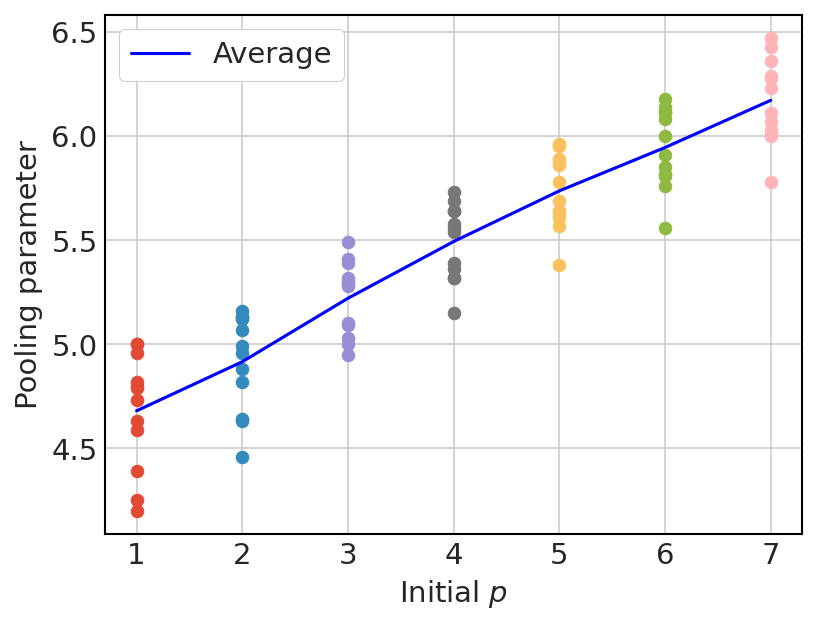}
        \caption{GGeM pooling}
    \end{subfigure}%
\caption{\textbf{Trained pooling parameters based on initial $p$.} We visualize trained pooling parameters by different initial $p$ on GeM and GGeM pooling. GeM pooling has a single shared parameter, while GGeM pooling has 12 pooling parameters for ViT-B/16.}
\label{fig:supp_p}
\end{figure*}

\paragraph{Impact of Initial $p$.}
We differentiate initial $p$ for GeM with fixed $p$, GeM with trainable $p$, and GGeM with trainable $p$.
As shown in Fig.~\ref{fig:power}, the performances are affected by the initial $p$.
GGeM pooling consistently outperforms GeM with fixed $p$ and GeM with trainable $p$, except for the case of $p=7$.
GeM with fixed $p$ and trainable $p$ seems to have similar performance patterns.
In all three cases, performances are increased by increasing the initial $p$ and dropped after the optimal $p$.
The optimal value of the initial $p$ can vary by the models and the tasks, but the range of 3 to 5 shows stable performance for GGeM.

\paragraph{Impact of Number of Groups.}
Fig.~\ref{fig:group} shows the influence of the number of groups $G$ by different initial power $p$. $G=1$ is the GeM pooling with a single pooling parameter for all channel dimensions, while $G=768$ indicates using different pooling parameters for each channel dimension.
The performances for initial $p=3$ and $p=5$ increase until the $G$ equals to the number of heads ($G=12$ for ViT-B/16), then it drops until $G=768$.
In other words, the optimal $G$ is the number of heads in ViT.
The results are linked to the analysis in Sec.~\ref{sec:analysis}.
Because of the MHA architecture, using multiple $p$ within a head can be harmful by the different degrees of gradient concentration for the different dimensions of the head, while using a single $p$ for all channel dimensions cannot be beneficial for achieving a small inter-head similarity.

\paragraph{Trained Pooling Parameters}
As the pooling parameter $p_i$ is trainable, the trained value will be different from the initial value.
We visualize how the pooling parameter changes with different initial values for GeM and GGeM pooling.
As shown in Fig.~\ref{fig:supp_p}, the trained pooling parameter gets higher with the increase of the initial value, and every trained pooling parameter converges between 4 and 6.5.
When the initial value is 1, it increases until around 5, while the initial value of 7 decreases until around 6 to 6.5.
This indicates that there is a certain convergence range (\ie, 4 to 6.5).
The 12 pooling parameters of GGeM are distributed within a range of 1.

\subsubsection{Scratch Training and Fine-tuning}
In Tab.~\ref{table:classification}, we compare the pooling strategies based on the scratch training and fine-tuning on pre-trained models by Self-Supervised Learning (SSL), including MAE~\cite{he2021masked}, BeiT~\cite{bao2021beit}, SimMIM~\cite{xie2021simmim} and Data2Vec~\cite{baevski2022data2vec}.
Based on the trained model with label supervision, we report top-1 validation accuracy and $k$-NN performance ($k$=12).
$k$-NN performance is for an auxiliary metric to see the performance of feature representation itself without a linear classification layer.
For scratch training, we follow the training recipe of~\cite{he2021masked} with small modifications for stable training and better performance than~\cite{dosovitskiy2021an,touvron2021deit}.
As general scratch training techniques use the class token for a feature representation, our class token baseline shows a decent performance compared to DeiT.
While average and GeM pooling outperform the class token, max pooling shows mixed results on ViT-S/16 and ViT-B/16.
Overall, GGeM shows the best performance among all, having performance boosts between 0.3\%p and 0.7\%p.

For fine-tuning experiments, we fine-tune ImageNet-1K with SSL pre-trained models, which use the average pooling for baseline.
We follow the same fine-tuning recipe for each work and report our reproduced score (``Average pooling'') and reported score (``Reported'').
Based on the reproduced experiment, we differentiate pooling strategies.
Max pooling shows similar performances to the baseline, while GeM pooling shows better performances.
In comparison with Average pooling, GGeM shows the minimum of 0.2\%p and the maximum of 0.4\%p performance improvement on ViT-S/16 and ViT-B/16.
The results show that GGeM is a superior pooling strategy in both scratch training and fine-tuning in image classification task.

\begin{table*}[t]
\vspace{3mm}
\begin{adjustbox}{width=0.8\textwidth,center}
\centering
\begin{tabular}{lcccccccc>{\columncolor{Light}}c>{\columncolor{Light}}cc}
\toprule
         & \multicolumn{2}{c}{Class} & \multicolumn{2}{c}{Average} & \multicolumn{2}{c}{Max} & \multicolumn{2}{c}{GeM} & \multicolumn{2}{>{\columncolor{Light}}c}{\textit{GGeM}} & Reported \\ \cline{2-12} 
Method   & Acc.        & $k$-NN        & Acc.        & $k$-NN       & Acc.       & $k$-NN       & Acc.       & $k$-NN       & Acc.        & $k$-NN       & Acc.  \\
\midrule
\textit{\textbf{ViT-S/16}} &             &             &             &            &            &            &            &            &             &            &       \\
Scratch  & 80.0        & 79.2        & 80.4        & 80.2       & 77.1       & 76.3       & 80.6       & 80.4       & \textbf{80.7}        & \textbf{80.5}       & 79.9$^\ddagger$  \\
\midrule
\textit{\textbf{ViT-B/16}} &             &             &             &            &            &            &            &            &             &            &       \\
Scratch  & 82.4        & 81.8        & 82.4        & 82.2       & 82.6       & 82.4       & 82.6       & 82.6       & \textbf{82.7}        & \textbf{82.7}       & 81.8$^\ddagger$  \\
MAE~\cite{he2021masked}      & -           & -           & 83.5        & 83.0       & 83.5       & 83.1       & 83.7       & 83.2       & \textbf{83.9}        & \textbf{83.3}       & 83.6  \\
BeiT$^\dagger$~\cite{bao2021beit}     & -           & -           & 83.6        & 83.4       & 83.5       & 83.1       & 83.7       & 83.5       & \textbf{83.8}        & \textbf{83.6}       & 83.7  \\
SimMIM~\cite{xie2021simmim}   & -           & -           & 83.7        & 83.5       & 83.6       & 83.1       & 83.9       & 83.5       & \textbf{84.0}        & \textbf{83.6}       & 83.8  \\
Data2Vec~\cite{baevski2022data2vec} & -           & -           & 84.0        & 83.6       & 83.7       & 83.2       & 84.1       & 83.7       & \textbf{84.3}        & \textbf{83.8}       & 84.2  \\
\midrule
\textit{\textbf{ViT-L/16}} &             &             &             &            &            &            &            &            &             &            &       \\
MAE~\cite{he2021masked}      & -           & -           & 85.8        & 85.7       & 85.7       & 85.6       & 85.8       & 85.8       & \textbf{86.0}        & \textbf{85.9}       & 85.9  \\
BeiT$^\dagger$~\cite{bao2021beit}     & -           & -           & 85.9        & 85.8       & 86.0       & 85.8       & 86.1       & 85.9       & \textbf{86.2}        & \textbf{86.0}       & 86.0  \\
Data2Vec~\cite{baevski2022data2vec} & -           & -           & 86.5        & 86.4       & 86.6       & 86.4       & 86.6       & 86.4       & \textbf{86.7}        & \textbf{86.5}       & 86.6  \\
\bottomrule
\end{tabular}
\end{adjustbox}
\caption{\textbf{Scratch training (Scratch) and fine-tuning with SSL models for ImageNet-1K classification.} Every SSL model are pre-trained with ImageNet-1K, except BeiT~\cite{bao2021beit} with $^\dagger$ used ImageNet-22K. We report top-1 accuracy (Acc.) and $k$-NN score based on trained model with label supervision. ``Reported'' indicates reported score in the original paper. $^\ddagger$ denotes scores from DeiT~\cite{touvron2021deit}.\vspace{-3mm}}
\label{table:classification}
\end{table*}
\begin{table}[t]
\begin{adjustbox}{width=1.0\columnwidth,center}
\centering
\begin{tabular}{lcccccccc}
\toprule
\multirow{2}{*}{Pooling} & \multicolumn{8}{c}{\# blocks fine-tuned}              \\ \cline{2-9} 
                         & 0    & 1    & 2    & 4    & 6    & 8    & 10   & 12   \\
\midrule
Class                    & \textbf{67.8} & 75.2 & 79.1 & 82.0 & 82.9 & 83.2 & 83.4 & 83.4 \\
Average                  & 66.6 & 77.5 & 80.4 & 82.2 & 82.9 & 83.3 & 83.4 & 83.5 \\
Max                      & 59.1 & 77.2 & 79.8 & 81.9 & 82.9 & 83.2 & 83.2 & 83.5 \\
GeM                      & 65.9 & 78.6 & 80.7 & 82.4 & 83.1 & 83.4 &  83.6 & 83.7 \\
\rowcolor{Light}\textit{GGeM}            & 65.9 & \textbf{79.0} & \textbf{80.8} & \textbf{82.5} & \textbf{83.2} & \textbf{83.5} & \textbf{83.7} & \textbf{83.9} \\
\bottomrule
\end{tabular}
\end{adjustbox}
\caption{\textbf{Linear probing and partial fine-tuning.} We fine-tune partial Transformer blocks with different poolings. Tuning 0 block is linear probing and tuning 12 blocks is full fine-tuning.}
\label{table:partial}
\end{table}

\subsubsection{Linear Probing and Partial Fine-tuning}
We conduct linear probing and partial fine-tuning experiments based on MAE pre-trained ViT-B/16 model.
For linear probing, we train a linear classifier on frozen blocks.
For partial fine-tuning, we fine-tune the last several layers based on other frozen blocks by following~\cite{he2021masked}.
For both GeM and GGeM, initial $p$ is set to 3.
Tab.~\ref{table:partial} shows the results.
Notably, all pooling methods are less linearly separable compared to the class token for linear probing.
However, pooling methods generally outperform for a few blocks of training (\ie, 1 to 4), which indicates the pooling methods perform well when non-linear layers are tuned.
For the number of blocks 6 to 12, class token shows similar performance with average and max pooling.
GeM and GGeM share similarly high performance, while GGeM shows the best performances for all partial fine-tuning cases.
The results demonstrate that GGeM can be used for competitive representations for non-linear layer tuning.

\subsubsection{Robustness}
We conduct an experiment to verify how robust pooling methods are to uncommon examples.
We quantitatively benchmark robustness in background change~\cite{xiao2020noise} with fine-tuned models on MAE.
Ideally, robust models can handle background variations and locate discriminative foreground parts.
With such motivation, the background change benchmark introduces ImageNet-9 (IN-9) dataset.
IN-9 includes 9 coarse-grained classes and 7 variants, which are generated by mixing images from different foregrounds and backgrounds.
Only-FG (O.F.), Mixed-Same (M.S.), Mixed-Rand (M.R.), and Mixed-Next (M.N.) are 4 variants with the original foreground and the modified background.
No-FG (N.F.), Only-BG-B (O.BB.), and Only-BG-T (O.BT.) are 3 variants with the masked foreground.
As demonstrated in Tab.~\ref{table:supp_robust}, GGeM shows the best performance in all variants with the highest mean performance.
The results signify that the training with GGeM pooling can result in a robust model for uncommon examples.

\begin{table}[t]
\begin{adjustbox}{width=1.0\columnwidth,center}
\centering
\begin{tabular}{lccccccccc}
\toprule
Pooling         & O.F.   & M.S.  & M.R.  & M.N.  & N.F.  & O.BB. & O.BT. & IN-9  & Mean \\
\midrule
Class    & 81.3 & 84.3 & 77.5 & 75.0 & 47.4 & 18.8 & 12.9 & 94.0 & 61.7   \\
Average  & 81.5 & 84.1 & 77.3 & 75.2 & 47.5 & 19.2 & 13.3 & 93.8 & 61.5   \\
Max      & 81.6 & 84.2 & 77.6 & 76.4 & 48.4 & 18.8 & 12.7 & 94.0 & 61.7   \\
GeM      & 81.5 & 84.6 & 78.0 & 76.7 & 47.9 & 19.3 & 13.3 & 93.9 & 61.9   \\
\rowcolor{Light}
\textit{GGeM}     & \textbf{82.0} & \textbf{85.1} & \textbf{78.2} & \textbf{76.8} & \textbf{48.6} & \textbf{19.7} & \textbf{13.5} & \textbf{94.3} & \textbf{62.0}   \\
\bottomrule
\end{tabular}
\end{adjustbox}
\caption{\textbf{Robustness evaluation of pre-trained models against background change.}}
\label{table:supp_robust}
\end{table}

\begin{table*}[t!]
    \centering
    \begin{subtable}[h]{0.75\textwidth}
        \begin{adjustbox}{width=1.0\textwidth,center}
        \centering
        \begin{tabular}{lcccccccccccccc}
        \toprule
              & \rot{ImageNet} & \rot{Cifar10} & \rot{Cifar100} & \rot{CLEVR-C} & \rot{DTD}  & \rot{EuroSAT} & \rot{FER2013} & \rot{Food101} & \rot{GTSRB} & \rot{MNIST} & \rot{RESISC45} & \rot{StanfordCars} & \rot{STL10} & \rot{Mean} \\
        \midrule
        Class & 11.8     & 42.4    & 16.2     & 8.8      & 6.8  & 14.6    & 15.1    & 8.0     & 6.4   & 6.3   & 19.1     & 0.8          & 69.7  & 17.4    \\
        Average& 12.9     & 44.9    & \textbf{19.4}     & 11.2     & 8.0  & 15.6    & 13.9    & 8.0     & 5.5   & \textbf{10.8}  & 19.7     & \textbf{1.1}          & 69.1  & 18.5    \\
        Max  & 12.6     & 42.5    & 16.4     & 13.0     & \textbf{10.0} & 18.9    & 13.1    & 8.2     & 6.6   & 4.4   & \textbf{21.7}     & 0.8          & 66.4  & 18.1    \\
        GeM   & \textbf{13.6}     & 48.4    & 16.5     & 12.2     & 7.9  & \textbf{19.8}    & 15.7    & 8.4     & \textbf{9.9}   & 6.8   & 19.2     & 0.8          & 69.5  & 19.1    \\
        \rowcolor{Light} \textit{GGeM}  & 13.0     & \textbf{49.6}    & 18.0     & \textbf{15.6}     & 8.5  & 13.7    & \textbf{21.0}    & \textbf{9.0}     & 8.2   & 10.0  & 20.4     & 0.9          & \textbf{71.5}  & \textbf{19.9}   \\
        \bottomrule
        \end{tabular}
        \end{adjustbox}
    \caption{Zero-shot classification}
    \vspace{2mm}
    \label{table:multimodal_classification}
    \end{subtable}
    \begin{subtable}[h]{0.75\textwidth}
        \begin{adjustbox}{width=1.0\textwidth,center}
        \centering
        \begin{tabular}{lcccccccccccc}
        \toprule
             & \multicolumn{6}{c}{Image-to-Text}                          & \multicolumn{6}{c}{Text-to-Image}                          \\
               & \multicolumn{3}{c}{Flickr30k} & \multicolumn{3}{c}{MSCOCO} & \multicolumn{3}{c}{Flickr30k} & \multicolumn{3}{c}{MSCOCO} \\
              & R@1     & R@5     & R@10   & R@1      & R@5      & R@10    & R@1     & R@5     & R@10   & R@1     & R@5      & R@10     \\
        \midrule
        Class & 19.8    & 42.5    & 53.9   & 8.0      & 23.2     & 33.4    & 13.5    & 31.5    & 41.4   & 6.7     & 19.3     & 27.5     \\
        Average& \textbf{20.4}    & 42.8    & 54.3   & 9.6      & 25.2     & 35.6    & \textbf{14.7}    & 31.9    & 41.3   & 7.2     & 19.8     & 28.1     \\
        Max  & 17.5    & 40.4    & 53.7   & 9.9      & 25.7     & 35.7    & 14.0    & 32.7    & 42.9   & 7.6     & 20.2     & 29.1     \\
        GeM   & 19.7    & \textbf{45.8}    & \textbf{57.3}   & 10.7     & 26.9     & 36.9    & 14.2    & 33.5    & 43.2   & 7.9     & 20.7     & 29.4     \\
        \rowcolor{Light} \textit{GGeM}  & \textbf{20.4}    & 45.3    & 56.7   & \textbf{11.5}     & \textbf{27.7}     & \textbf{37.9}    & 14.4    & \textbf{33.7}    & \textbf{44.0}   & \textbf{8.4}     & \textbf{21.3}     & \textbf{30.3}    \\
        \bottomrule 
        \end{tabular}
        \end{adjustbox}
    \caption{Zero-shot cross-modal retrieval}
    \label{table:multimodal_retrieval}
    \end{subtable}
\caption{\textbf{Multi-modal representation learning with different pooling methods.}}
\label{tab:multimodal}
\end{table*}
\begin{table}[t]
\begin{adjustbox}{width=1.0\columnwidth,center}
\centering
\begin{tabular}{lccccccccc}
\toprule
\multirow{2}{*}{Pooling} & \multicolumn{3}{c}{CUB200} & \multicolumn{3}{c}{Cars196} & \multicolumn{3}{c}{SOP} \\ \cline{2-10} 
                         & R@1     & RP      & mAP    & R@1     & RP      & mAP     & R@1    & RP     & mAP   \\
\midrule
Class                    & 71.9   & 42.4   & 31.7  & 92.1   & 46.8   & 37.9   & 82.7  & 60.5  & 57.6 \\
Average                  & 71.3   & 41.4   & 30.7  & 92.3   & 47.2   & 38.2   & 82.9  & 61.0  & 58.1 \\
Max                      & 74.6   & 43.6   & 33.1  & 93.2   & 49.1   & 40.6   & 83.0  & 61.2  & 58.3 \\
GeM                      & 75.0   & 44.1   & 34.1  & 93.8   & 50.6   & 42.5   & 83.7  & 62.0  & 59.2 \\
\rowcolor{Light}
\textit{GGeM}                     & \textbf{75.2}   & \textbf{45.0}   & \textbf{34.6}  & \textbf{94.1}   & \textbf{51.0}   & \textbf{43.0}   & \textbf{84.8}  & \textbf{62.1}  & \textbf{59.3} \\
\bottomrule
\end{tabular}
\end{adjustbox}
\caption{\textbf{Image retrieval with different pooling methods.}}
\label{table:retrieval}
\end{table}

\subsection{Image Retrieval}
\label{sec:retrieval}
Pooling strategies are actively studied in image retrieval tasks to find the best spatial information aggregator.
We fine-tune MAE~\cite{he2021masked} pre-trained ViT-B/16 models on CUB200~\cite{wah2011cub}, Cars196~\cite{krause2013cars}, and Stanford Online Products (SOP)~\cite{oh2016deep} with Norm-softmax loss~\cite{wang2018additive,wang2017normface}.
For the fair comparison, we use three different metrics: Recall@1, mean Average Precision (mAP), and R-Precision (RP) as recall-based metrics can lead to an unreliable conclusion compared to precision-based metrics~\cite{musgrave2020metric,chun2021pcme,chun2022eccv_caption}.
We use initial $p=3$ for both GeM and GGeM.
As shown in Tab.~\ref{table:retrieval}, Max, GeM and GGeM pooling show higher performances compared to class token and average pooling, while GGeM pooling achieves the best for every benchmark.
The results support the superiority of GGeM in aggregating per-token information for better feature representation.

\subsection{Multi-modal Representation Learning}
\label{sec:multimodal}
Large-scale vision-language pre-training models, such as CLIP~\cite{radford2021clip} and ALIGN~\cite{jia2021scaling}, has demonstrated success over various downstream tasks.
ViT-based CLIP models use the class token as image representation in the visual encoder and $[EOS]$ (end of a sentence) token as text representation in the text encoder.
Pooling strategies can be unsuitable for the text encoder because text inputs have different lengths, and the remaining tokens are filled with padding $[PAD]$.
Thus, we differentiate pooling strategies in the visual encoder (ViT-B/32) and trained the multi-modal model with the Conceptual Captions (CC3M) dataset~\cite{sharma2018conceptual}.
We use following benchmark datasets for evaluating zero-shot classification: ImageNet~\cite{deng2009imagenet}, Cifar10~\cite{krizhevsky2009learning}, Cifar100~\cite{krizhevsky2009learning}, CLEVR Counts (CLEVR-C)~\cite{johnson2017clevr}, Describable Textures Dataset (DTD)~\cite{cimpoi14describing}, EuroSAT~\cite{helber2019eurosat}, FER2013~\cite{goodfellow2013challenges}, Food101~\cite{bossard2014food}, GTSRB~\cite{stallkamp2012man}, MNIST~\cite{lecun1998mnist}, RESISC45~\cite{cheng2017remote}, StanfordCars~\cite{krause20133d}, STL10~\cite{coates2011analysis}.
For zero-shot cross-modal retrieval, we use Flickr30k~\cite{plummer2015flickr30k} and MSCOCO~\cite{lin2014microsoft} benchmark datasets.
The initial $p$ is set to 3 for both GeM and GGeM.
The results are given in Tab.~\ref{tab:multimodal}.
In the zero-shot classification tasks, the best scores for each benchmark dataset vary by the choice of pooling strategies, but GGeM shows the best performance on average among all.
Moreover, GGeM shows mixed performances in MSCOCO, while showing the best performance in Flickr30k.
The results signify the superiority of GGeM for training large-scale vision-language pre-training tasks.
\section{Conclusion}
\label{sec:conclusion}
In this paper, we have introduced GGeM pooling, which considers channel-wise differences in the activation maps.
GGeM aggregates per-token information of ViT by dividing channels into groups and computing GeM pooling with shared pooling parameter per group.
Based on our analysis, grouping channels as the number of heads in ViT can induce heads to learn different characteristics from each other.
We have shown that GGeM outperforms existing pooling strategies for scratch training and fine-tuning in image classification while showing potential to be used for other tasks such as image retrieval and multi-modal representation learning.
The limitation of our work includes GGeM can work differently on ViT variants, which use a different number of heads within a model (\ie, \cite{liu2021swin}).
Moreover, GGeM is limited to consider channel-wise difference within the heads because of the shared pooling parameter per group.
Exploring GGeM on other ViT variants and considering channel-wise difference even within the heads will be interesting future research directions.

{\small
\bibliographystyle{ieee_fullname}
\bibliography{references}
}

\clearpage



\section*{Supplementary material}

# Group Generalized Mean Pooling for Vision Transformer

Supplementary Material

## Contents

## A. Analysis Details

### A.1. Additional Results on Impact of Grouping

To analyze the impact of grouping in GGeM, we compute the similarity of heads in ViT. We use Centered Kernel Alignment (CKA) similarity [4] which is designed to compare representations within and across networks quantitatively. CKA similarity takes representations of two layers, $\mathbf{X} \in \mathbb{R}^{(s \times n_1)}$ and $\mathbf{Y} \in \mathbb{R}^{(s \times n_2)}$, where $s$ and $n_i$ are number of samples and neurons, respectively. Given $\mathbf{G} = \mathbf{X}\mathbf{X}^T$ and $\mathbf{H} = \mathbf{Y}\mathbf{Y}^T$ as the Gram matrices for the two layers, CKA similarity is defined by:

$$CKA(\mathbf{G}, \mathbf{H}) = \frac{HSIC(\mathbf{G}, \mathbf{H})}{\sqrt{HSIC(\mathbf{G}, \mathbf{G}) HSIC(\mathbf{H}, \mathbf{H})}}, \quad \text{(i)}$$

where $HSIC$ is the Hilbert-Schmidt independence criterion [2]. In addition to the Fig. 4, we further visualize the mean of CKA similarities between heads in Transformer blocks 7 to 12. As shown in Fig. A, GGeM shows smaller inter-head similarity compared to other pooling methods in Transformer blocks 9 to 12. The closer the Transformer block to the pooling layer, the larger such pattern stands out. The results demonstrate that GGeM is helpful for inducing the heads to learn different representation subspaces.

### A.2. Additional Results on Attention Head Mean Distance

We analyze how pooling strategies affect ViT models by comparing attention head mean distances. In addition to the Fig. 5, we further compute attention head mean distances of Transformer blocks 7 to 12. As shown in Fig. B, the mean distances increase by the pooling parameter $p$ increases. Such a phenomenon gets more noticeable when the Transformer block is higher and closer to the pooling layer. The results show that the larger $p$ increases the number of heads attending on the global information.

## B. Implementation Details

### B.1. Pseudo Code

The proposed GGeM is a simple algorithm that only a few lines of code are necessary for implementation. We include PyTorch-like pseudo-code of GGeM in Algorithm 1. When `groups= 1`, the algorithm works as GeM pooling. GGeM with `groups= 1` and fixed `p_params= 1` will work as Average pooling. We set `groups` as the number of heads in each ViT architecture and have a shared pooling parameter per group. The `clamp` function is for ensuring every elements are larger than zero.

### B.2. Experimental Setting

The default setting for ImageNet classification experiments is in Tab. A. For scratch training, we follow the training setting of [3] with modification of the learning rate for a better baseline. The exponential moving average is used

---

**Algorithm 1** GGeM: PyTorch-like Pseudo-code

```python
# x: output tensor of the last Transformer block (
    batch size x tokens x dimension)
# p_params: pooling parameters, the number of
    pooling parameters is the same with groups
# groups: number of groups
# eps: minimum value for clamping
def GGeM(x):
    x = x[:, 1:, :] # Remove class token

    b, t, d = x.shape
    e = d // groups
    x = x.reshape((b, t, e, groups))

    x = x.clamp(min=eps).pow(p_params)
    x = x.mean(dim=1)
    x = x.pow(1./p_params)

    x = x.reshape((b, d))
    return x
```

**Notes**: `clamp` is a function to clamp all elements into the range [min, max].

---

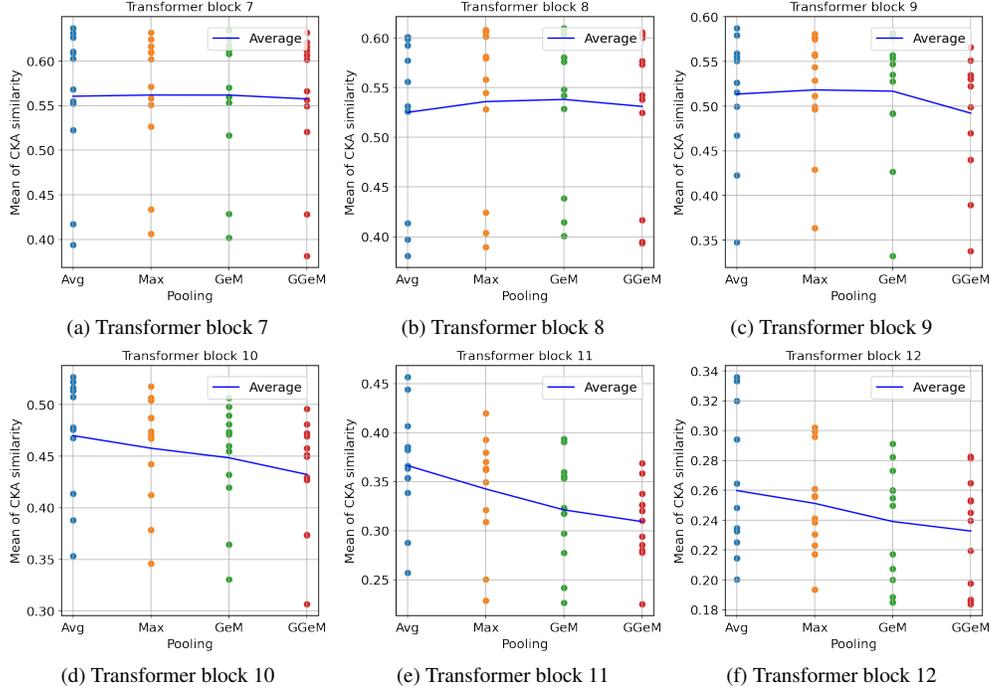

Figure A. **Mean of CKA similarities between heads of MHA.** We visualize Transformer blocks 7 to 12 of ViT-B/16. Each dot denotes mean of CKA similarities between the query head and other heads. The blue line denotes average of all CKA similarities between heads.

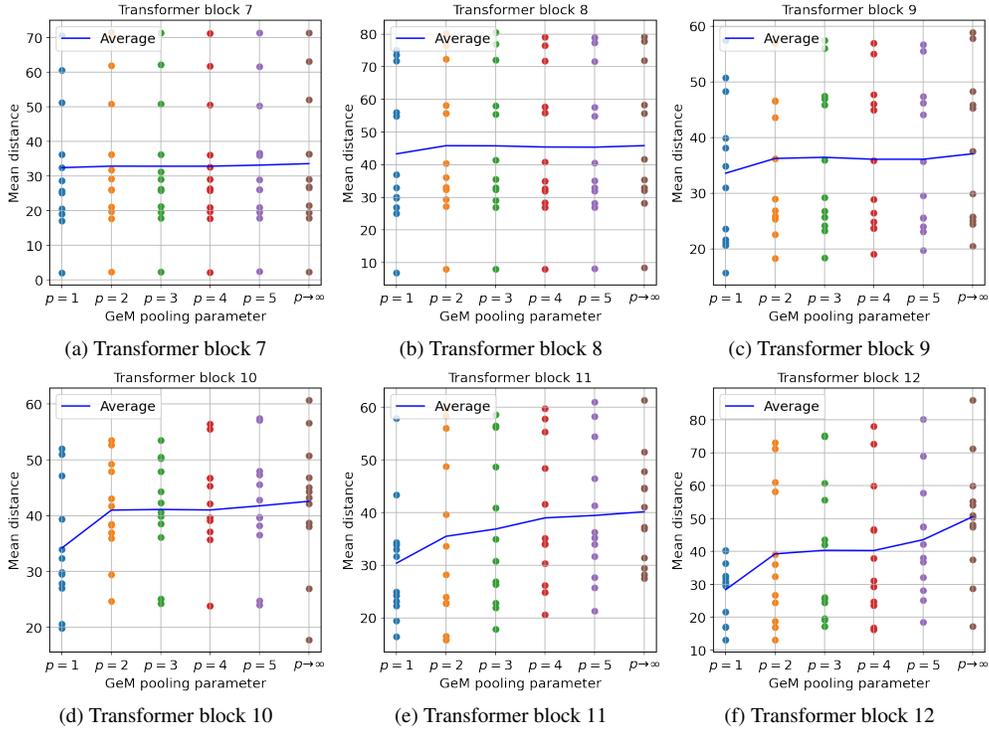

Figure B. **Attention head mean distances with different GeM pooling parameters.** We visualize Transformer blocks 7 to 12 of ViT-B/16. Higher mean distance indicates more globally attending heads. $p = 1$ and $p \to \infty$ are average and max pooling, respectively.

| Config | Scratch | MAE | BeiT | SimMIM | Data2Vec | Linear probing |
|---|---|---|---|---|---|---|
| Initial p | 5 | 5 (B), 2 (L) | 5 | 4 | 3 | 3 |
| Optimizer | AdamW | AdamW | AdamW | AdamW | AdamW | LARS |
| Learning rate | 8e-3 | 2e-3 | 4e-3 | 5e-3 | 4e-3 (B), 5e-3 (L) | 0.64 |
| Weight decay | 0.3 | 0.05 | 0.05 | 0.05 | 0 (B), 0.05 (L) | 0 |
| Optimizer momentum | $\beta_1, \beta_2 = 0.9, 0.95$ | $\beta_1, \beta_2 = 0.9, 0.999$ | $\beta_1, \beta_2 = 0.9, 0.999$ | $\beta_1, \beta_2 = 0.9, 0.999$ | $\beta_1, \beta_2 = 0.9, 0.999$ | 0.9 |
| Layer decay | 0.75 | 0.65 (B), 0.75 (L) | 0.65 (B), 0.75 (L) | 0.65 | 0.65 | - |
| Batch size | 4096 | 1024 | 1024 | 2048 | 1024 | 16384 |
| LR schedule | cosine decay | cosine decay | cosine decay | cosine decay | cosine decay | cosine decay |
| Warmup epochs | 20 | 5 (B), 10 (L) | 20 (B), 5 (L) | 20 | 10 (B), 5 (L) | 10 |
| Traning epochs | 300 | 100 (B), 50 (L) | 100 (B), 50 (L) | 100 | 100 (B), 50 (L) | 90 |
| Augmentation | RandAug (9, 0.5) | RandAug (9, 0.5) | RandAug (9, 0.5) | RandAug (9, 0.5) | RandAug (9, 0.5) | RandomResizedCrop |
| Label smoothing | 0.1 | 0.1 | 0.1 | 0.1 | 0.1 | - |
| Mixup | 0.8 | 0.8 | 0.8 | 0.8 | 0.8 | - |
| Cutmix | 1.0 | 1.0 | 1.0 | 1.0 | 1 | - |
| Drop path | 0.1 | 0.1 (B), 0.2 (L) | 0.1 (B), 0.2 (L) | 0.1 | 0.2 (B), 0.25 (L) | - |
| Exp. moving average | 0.9999 | - | - | - | - | - |

Table A. **Hyper-parameter setting for ImageNet-1K classification.**

| Config | Value |
|---|---|
| Initial p | 3 |
| Model | ViT-B/16 |
| Optimizer | AdamW |
| Learning rate | 2e-3 |
| Weight decay | 0.05 |
| Optimizer momentum | $\beta_1, \beta_2 = 0.9, 0.999$ |
| Layer-wise LR decay | 0.65 |
| Batch size | 128 |
| Learning rate schedule | cosine decay |
| Warmup epochs | 5 |
| Traning epochs | 100 (Cars, CUB), 50 (SOP) |
| Augmentation | RandAug (9, 0.5) |
| Drop path | 0.1 |
| Base learning rate | 4e-3 |
| Proxy LR scale | 100 |

(a) Image retrieval setting

| Config | Value |
|---|---|
| Initial p | 3 |
| Model | ViT-B/32 |
| Optimizer | AdamW |
| Learning rate | 1e-3 |
| Weight decay | 0.2 |
| Optimizer momentum | $\beta_1, \beta_2 = 0.9, 0.98$ |
| Batch size | 4080 |
| Learning rate schedule | cosine decay |
| Warmup steps | 3500 |
| Traning epochs | 25 |
| Augmentation | RandomResizedCrop |

(b) Multi-modal representation learning setting

Table B. **Hyper-parameter settings for image retrieval and multi-modal representation learning.**

only in scratch training. For fine-tuning, we follow the same training recipe of each corresponding work (MAE, BeiT, SimMIM, and Data2Vec) based on each pre-trained model and sweep over different initial $p$ (1 to 5) to find the best performance for GeM and GGeM. We conduct the linear probing experiment by following [1], while we use the same training setting of MAE for partial fine-tuning.

The hyper-parameter settings of image retrieval and multi-modal representation learning are in Tab. B. We conduct the image retrieval experiment based on MAE fine-tuning code. We use Norm-softmax loss [9,10] as a baseline objective function while scaling the learning rate of proxy (proxy LR scale) with 100 for better baseline performance. For multi-modal representation learning, we use the same model architecture of CLIP [7] and the experimental setting of SLIP [5]. We use ViT-B/32 image encoder and the Transformer-based text encoder [8] with 49,152 vocab size of byte pair encoding (BPE) tokenizer. The max sequence length was clipped by 76 by following CLIP. We implement all models using the PyTorch framework [6] and use four 8 × A100-80GB servers for experiments.

## C. Additional Experiments

### C.1. Visualization

We visualize Class Activation Map (CAM) [12] to see how models trained with different pooling methods make specific decisions based on the input image. For the CAM method, Score-CAM [11] is used, which is a variant of CAM which removes dependence on gradients. We use ViT-B/16 fine-tuned models based on MAE, and images are randomly selected from the ImageNet validation set. The results are shown in Fig. C.

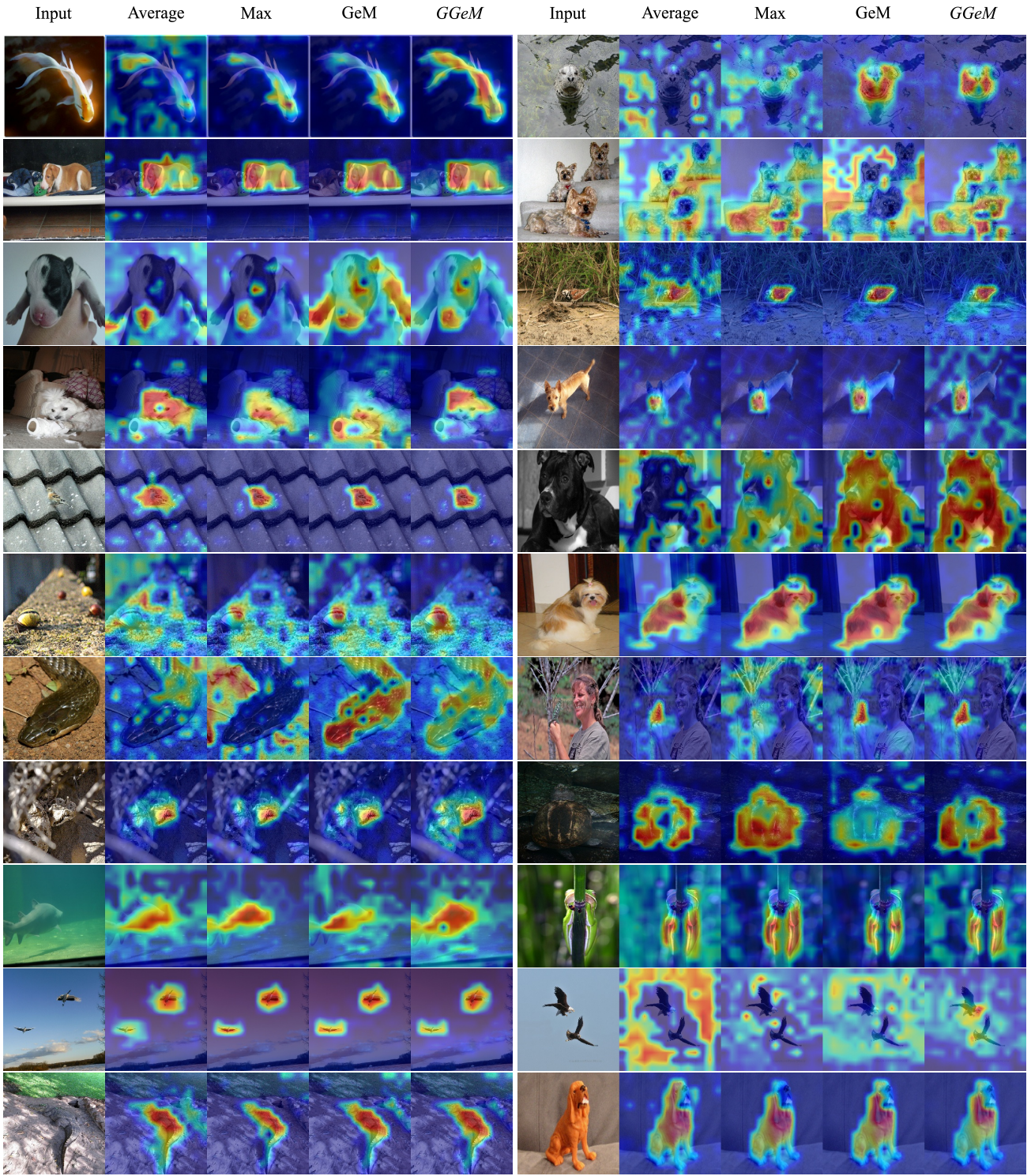

Figure C. **Score-CAM visualization with different pooling methods.**

5



\end{document}